# Stitching the Story: Creating Panoramic Incident Summaries from Body-Worn Footage

Dor Cohen
*Software Engineering, Afeka Academic College of Engineering*
Tel Aviv Israel

Inga Efrosman
*Software Engineering, Afeka Academic College of Engineering*
Tel Aviv Israel

Yehudit Aperstein
*Intelligent Systems, Afeka Academic College of Engineering*
Tel Aviv Israel

Alexander Apartsin
*School Of Computer Science, Faculty Of Sciences*
Holon Institute of Technology

*Abstract*— First responders widely adopt body-worn cameras to document incident scenes and support post-event analysis. However, reviewing lengthy video footage is impractical in time-critical situations. Effective situational awareness demands a concise visual summary that can be quickly interpreted. This work presents a computer vision pipeline that transforms body-camera footage into informative panoramic images summarizing the incident scene. Our method leverages monocular Simultaneous Localization and Mapping (SLAM) to estimate camera trajectories and reconstruct the spatial layout of the environment. Key viewpoints are identified by clustering camera poses along the trajectory, and representative frames from each cluster are selected. These frames are fused into spatially coherent panoramic images using multi-frame stitching techniques. The resulting summaries enable rapid understanding of complex environments and facilitate efficient decision-making and incident review.

Keywords—Body-worn cameras; Situational awareness; Video summarization; Panoramic image stitching; Simultaneous Localization and Mapping (SLAM); Camera pose estimation

## I. Introduction

Body-worn cameras have become indispensable tools for first responders across various domains, including law enforcement, firefighting, and emergency medical services. These cameras capture first-person perspectives of dynamic and often chaotic environments, providing valuable evidence for post-incident analysis, training, and accountability. However, video data's sheer volume and temporal nature present a significant bottleneck: reviewing footage in real time or post hoc can be prohibitively time-consuming and cognitively demanding, especially in high-stakes scenarios where rapid decision-making is essential.

In such contexts, static visual summaries offer a compelling alternative. Unlike full-length videos, static summaries such as curated sets of panoramic images can provide a comprehensive situational overview at a glance. These images enable investigators, supervisors, or additional responding units to quickly orient themselves to the incident scene's spatial structure and key elements without wading through hours of video. Moreover, visual summaries can be archived and retrieved efficiently, making them suitable for documentation, evidence sharing, and after-action reviews.

Recent advances in computer vision and AI provide powerful tools for automatically generating such summaries from video. Techniques like Simultaneous Localization and Mapping (SLAM), deep learning-based frame selection, and image stitching make reconstructing scenes, identifying critical viewpoints, and producing spatially accurate panoramic representations feasible. By applying these techniques to body-camera footage, we can build systems that intelligently distil hours of egocentric video into compact, high-value visual artifacts.

This paper proposes a novel pipeline that leverages monocular SLAM to estimate the camera trajectory, clusters the path into spatial segments, and selects representative frames within each cluster to generate informative panoramic images. Our method balances the need for spatial accuracy, semantic relevance, and visual clarity, providing a scalable solution for incident scene summarization.

## II. Literature review

### A. Situation awareness and body-worn cameras

Situational awareness (SA), defined as the ability to perceive, comprehend, and project environmental elements, is critical for first responders operating in dynamic, high-risk environments (Endsley, 1995). Body-worn cameras (BWCs) have emerged as tools to enhance SA by documenting incidents and providing post-hoc accountability, yet their real-time utility remains debated. This review synthesizes academic research on the role of BWCs in SA, focusing on empirical outcomes, technological integration, and ethical implications. Endsley's (1995) three-level model of SA (perception, comprehension, projection) underpins research on first responders' decision-making. Cognitive overload in chaotic environments often degrades SA, necessitating tools like BWCs to augment perception (Braga et al., 2018). However, BWCs may inadvertently distract officers during critical moments, undermining SA (Smith & Williams, 2017). Mixed evidence characterizes BWCs' impact. A randomized controlled trial (RCT) in Las Vegas linked BWCs to a 17% reduction in civilian complaints and 10% fewer use-of-force incidents, suggesting improved accountability and SA-driven de-escalation (Braga et al., 2018). Conversely, officers in some jurisdictions experienced a 37% increase in assaults, attributed to hesitancy in assertive tactics due to surveillance awareness (Ariel et al., 2018). Meta-analyses highlight contextual variability, with effects mediated by community trust and policy implementation (National Institute of Justice, 2022).

AI-enhanced BWCs, such as facial recognition systems, improved suspect identification rates by 22% in field trials, directly aiding SA (Park & Pang, 2019). Real-time streaming to command centers enables remote SA support during crises (White & Coldren, 2017). However, technical barriers persist, including data storage costs (3.75–7.55 GB/hour for HD video) and connectivity gaps in rural areas (National Institute of Justice, 2022).

Ethical concerns remain salient. BWCs' "observer effect" promotes procedural adherence among officers but risks exacerbating community distrust, particularly in marginalized groups (Smith & Williams, 2017). In healthcare settings, BWCs in mental health wards raise privacy concerns despite potential de-escalation benefits (Royal College of Nursing, 2021).

### B. Computer vision techniques for BWC footage processing and summarization

Computer vision and artificial intelligence (AI) techniques have been explored to address these challenges by enhancing SA through automated video summarization. For example, SLAM has been applied to BWC footage to reconstruct 3D scene layouts, enabling spatial awareness of complex environments (Chen, Zhang, & Li, 2021). Additionally, deep learning-based object detection and scene segmentation have been used to identify critical elements within footage, such as persons of interest or hazardous objects, thereby improving comprehension of incident scenes (Patel, Shah, & Gupta, 2022). Panoramic scene reconstruction from BWC footage enables spatially coherent summaries of incident environments. Monocular SLAM algorithms estimate camera trajectories and 3D environmental layouts, while clustering techniques identify key viewpoints for stitching representative frames into panoramas (Park & Pang, 2019). Multi-frame stitching algorithms, such as graph-based optimization, minimize distortions caused by erratic camera motion, a common challenge in egocentric video (Zaragoza et al., 2014). These panoramas aid investigators in visualizing spatial relationships during critical incidents, such as suspect positioning or hazard distribution (White & Coldren, 2017).

However, existing approaches have limitations. Automated summarization methods often prioritize visual fidelity over semantic relevance, resulting in summaries that may omit critical contextual details (Lee, Kim, & Park, 2022). These gaps highlight the need for robust pipelines that effectively support SA and balance spatial accuracy, semantic prioritization, and computational efficiency.

### C. Simultaneous Localization and Mapping and Parallel Tracking and Mapping

Simultaneous Localization and Mapping (SLAM) is a cornerstone of autonomous navigation and environmental reconstruction, enabling systems to map unknown environments while estimating their position. Early SLAM systems relied on probabilistic methods (Thrun et al., 2005), but visual SLAM (vSLAM) emerged as a cost-effective alternative using camera data. PTAM (Parallel Tracking and Mapping) was a pivotal advancement that separated camera tracking and mapping into parallel threads, enabling real-time performance (Klein & Murray, 2007). PTAM's feature-based approach, using keypoints like ORB (Rublee et al., 2011), laid the groundwork for modern SLAM systems. However, challenges such as scale ambiguity in monocular SLAM (Mur-Artal et al., 2015) and dynamic scene handling persist. Recent work integrates deep learning for semantic understanding (Cadena et al., 2016), enhancing robustness in complex environments.

### D. Point Cloud Clustering and Dominant Set Clustering

Point cloud segmentation is vital for interpreting 3D environments. Traditional methods like DBSCAN (Ester et al., 1996) group points by density but struggle with variable densities. Graph-based approaches, such as Normalized Cuts (Shi & Malik, 2000), use spatial relationships but face scalability issues. Dominant Set Clustering (Pavan & Pelillo, 2003), a graph-theoretic method, identifies clusters by maximizing intra-cluster similarity, offering advantages in semantic segmentation. Recent deep learning models like PointNet (Qi et al., 2017) directly process unstructured point clouds, achieving state-of-the-art results but requiring significant computational resources.

### E. Image Stitching Techniques

Image stitching combines overlapping images into panoramas, relying on feature detection and alignment. Early techniques used SIFT (Lowe, 2004) for keypoint matching, while modern methods employ deep learning for robustness to lighting variations (Brown & Lowe, 2007). Direct methods minimize photometric error but are sensitive to illumination changes (Zaragoza et al., 2014). Seam removal techniques, such as multi-resolution blending (Burt & Adelson, 1983) and graph-cut optimization (Kwatra et al., 2003), enhance visual coherence. Real-time stitching pipelines (Xu et al., 2020) are critical for applications like emergency response, where rapid summarization is essential.

### F. Integration of Techniques

Modern pipelines combine SLAM, clustering, and stitching for holistic scene understanding. Semantic SLAM systems (McCormac et al., 2018) use clustered point clouds to identify objects, improving navigation. For summarization, SLAM-derived trajectories guide keyframe selection (Mur-Artal & Tardós, 2017), enabling spatially coherent panoramas. These integrations highlight the synergy between geometric reconstruction and semantic analysis.

## III. SUMARIZATION PIPELINE

The proposed system processes body-worn camera footage to generate a set of informative panoramic images summarizing the spatial layout of incident scenes (Fig.1). It consists of three main conceptual and algorithmic phases.

The first phase performs keyframe selection and camera pose estimation using monocular SLAM techniques. The goal is to sparsely sample the video by identifying informative frames and estimating their positions and orientations in space. This is achieved through visual odometry, where motion is inferred from tracking visual features across frames. Keyframes are selected based on significant camera motion or scene change, and each keyframe is associated with a 6-degree-of-freedom pose representing its location and orientation in the reconstructed map. A sparse 3D point cloud is also created to support localization. This process is implemented using GPTAM, an adaptation of the PTAM system optimized for offline video processing. Unlike the original PTAM, which is designed for real-time augmented reality, GPTAM enables processing of pre-recorded footage, making it suitable for post-hoc summarization tasks. It separates tracking and mapping into parallel threads and uses OpenCV for feature detection, matching, and pose estimation. The modified library supports video file input and exports frame-wise pose estimates, enabling downstream spatial analysis.

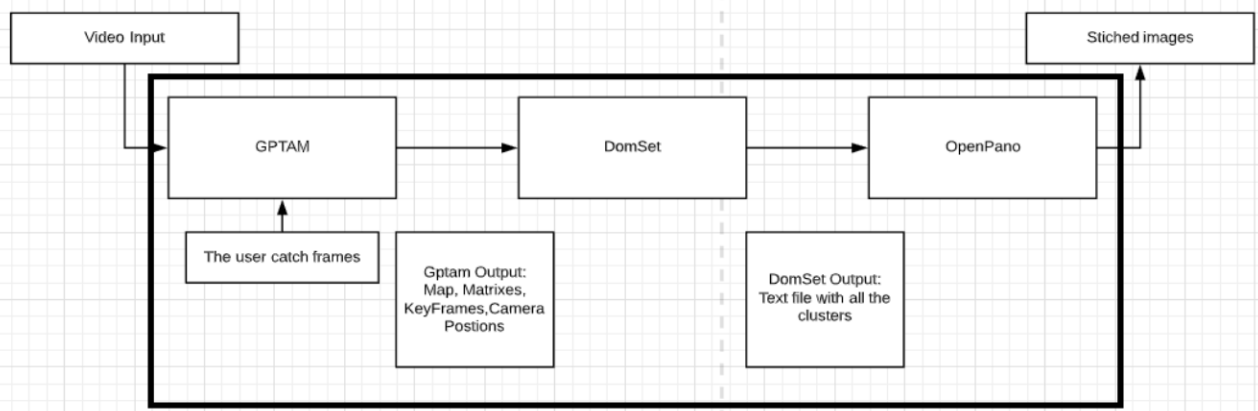

*Figure 1*: System Diagram

The second phase clusters the extracted keyframes into viewpoint groups using the Dominant Set clustering algorithm. A similarity graph is constructed in which each node represents a keyframe, and edge weights represent spatial or directional similarity between camera poses. Dominant Set clustering identifies maximally cohesive subsets of internally similar and externally dissimilar nodes, functioning like generalized cliques in edge-weighted graphs. The method does not require predefining the number of clusters and solves a quadratic optimization problem over the graph's affinity matrix. This is done using replicator dynamics, a technique derived from evolutionary game theory. The output is a set of clusters of spatially coherent keyframes, each representing a scene region suitable for panoramic stitching. The algorithm is particularly suitable for non-convex and irregular data distributions, such as those found in camera trajectories.

The third phase constructs panoramic images from the clustered keyframes. Each cluster creates a panoramic image through a pipeline of pairwise image alignment, global warping, and blending. Feature-based methods, such as SIFT or ORB, match keypoints across overlapping images, and transformations, such as homographies, are estimated to align them. Images are then projected into a shared panoramic coordinate system, cylindrical or spherical space, to correct for perspective and motion distortions. Seam optimization and multiband blending are applied to minimize visible stitching artifacts. This process produces a seamless panoramic image summarizing a portion of the original scene. The panoramic stitching uses the OpenPano toolkit, an open-source system that automates feature matching, alignment, and compositing of panoramic mosaics. OpenPano is implemented in C++ and supports efficient batch processing of image sets.

## IV. RESULTS

We conducted experiments using multiple camera video sequences to evaluate the proposed system for incident scene summarization. The primary objective was to assess the ability of the system to produce compact, informative panoramic images that capture the spatial layout and visual context of key regions along the responder's trajectory (Fig.2).

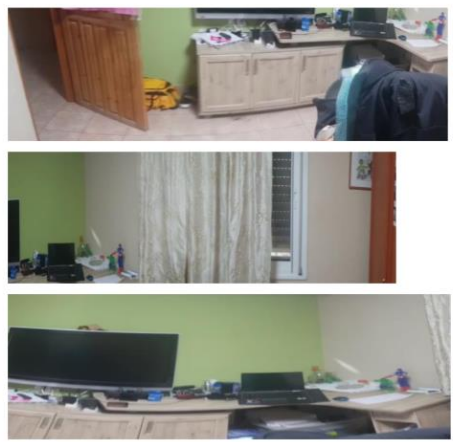

*Figure 2: Generated panoramic images from indoor camera footage*

The system was applied to each video in three phases: (1) keyframe extraction and pose estimation using GPTAM, (2) viewpoint clustering via Dominant Set clustering, and (3) panoramic image stitching within each cluster using the OpenPano framework. The clustering process organized camera poses into spatially coherent viewpoint groups, as illustrated in Fig. 3.

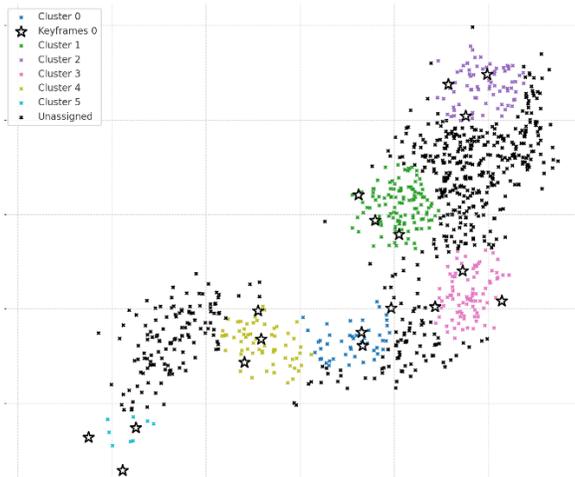

*Figure 3: Clustering viewpoints and keyframe locations*

## V. Conclusions and Future Work

This paper introduced a novel system for summarizing incident scenes from body-worn camera footage by generating spatially coherent panoramic images. The system operates in three conceptual phases: (1) extraction of keyframes and associated camera poses using an offline adaptation of monocular SLAM (GPTAM), (2) clustering of keyframes using the Dominant Set algorithm to group spatially consistent viewpoints, and (3) panoramic stitching within each cluster using feature-based mosaicking techniques implemented via OpenPano. The resulting image summaries offer a compact, structured visual representation of complex scenes, enabling rapid situational awareness without reviewing complete video sequences.

Future work should focus on quantitative and qualitative evaluation of the resulting summaries. Quantitatively, spatial coverage, stitching accuracy, redundancy reduction, and viewpoint distinctiveness could be used to assess summary fidelity. Qualitatively, structured user studies with first responders and investigators could evaluate interpretability, usefulness, and impact on decision-making and situational awareness. These evaluations will be essential to validate the operational value of the proposed system and to drive further research on real-time summarization, active summarization strategies, and integration with downstream incident analysis workflows.